# The Pragmatic Turn in Explainable Artificial Intelligence (XAI)

**Andrés Páez**
Universidad de los Andes
apaez@uniandes.edu.co

ABSTRACT

In this paper I argue that the search for explainable models and interpretable decisions in AI must be reformulated in terms of the broader project of offering a pragmatic and naturalistic account of understanding in AI. Intuitively, the purpose of providing an explanation of a model or a decision is to make it understandable to its stakeholders. But without a previous grasp of what it means to say that an agent *understands* a model or a decision, the explanatory strategies will lack a well-defined goal. Aside from providing a clearer objective for XAI, focusing on understanding also allows us to relax the factivity condition on explanation, which is impossible to fulfill in many machine learning models, and to focus instead on the pragmatic conditions that determine the best fit between a model and the methods and devices deployed to understand it. After an examination of the different types of understanding discussed in the philosophical and psychological literature, I conclude that interpretative or approximation models not only provide the best way to achieve the objectual understanding of a machine learning model, but are also a necessary condition to achieve post-hoc interpretability. This conclusion is partly based on the shortcomings of the purely functionalist approach to post-hoc interpretability that seems to be predominant in most recent literature.

## 1. Introduction

The main goal of Explainable Artificial Intelligence (XAI) has been variously described as a search for explainability, transparency and interpretability, for ways of validating the decision process of an opaque AI system and generating trust in the model and its predictive performance.[1] All of these goals remain underspecified in the literature and there are numerous proposals about which attributes make models interpretable. Instead of analyzing these goals and proposals piecemeal, the main contention of this paper is that the search for explainable, interpretable, trustworthy models and decisions[2] in AI must be reformulated in terms of the broader project of offering an account of *understanding* in AI. Intuitively, the purpose of providing an explanation or an interpretation of a model or a decision is to make it understandable or comprehensible to its stakeholders. But without a previous grasp of what it means to say that a human agent *understands* a decision or a model, the explanatory or interpretative strategies will lack a well-defined theoretical and practical goal. This paper provides a characterization of the theoretical goal of XAI by offering an analysis of human understanding in the context of machine learning in general, and of black box models in particular.

---

[1] For a survey of recent characterizations of the goals of XAI, see Lipton (2016), Doshi-Velez & Kim (2017), Samek et al. (2017), and Gilpin et al. (2019).

In recent years, there has been an increased interest in the notion of understanding among epistemologists (Pritchard 2014, Grimm 2018) and philosophers of science (de Regt et al. 2009). The interest in this notion has several sources. In epistemology, several authors realized that the conceptual analysis of understanding differs significantly from the traditional analysis of knowledge. In particular, unlike knowledge, understanding need not be factive: not all the information on the basis of which a phenomenon is understood must be true. Understanding is also an epistemic achievement that some authors regard as more valuable than mere knowledge. It also seems to be immune to Gettier cases, it is transparent to the epistemic agent, and it has internalist conditions of success. In sum, understanding and knowledge seem to be entirely different concepts and it is implausible to conceive the former simply as a species of the latter.[3]

In the philosophy of science, the first philosophers of explanation (Hempel 1965; Salmon 1984) regarded the understanding provided by a scientific explanation as a pragmatic and psychological by-product that falls beyond the ken of a proper philosophical theory. In their view, once we have developed an adequate account of explanation, any remaining questions regarding the notion of understanding can be addressed from a psychological perspective. A recent interest in the role of models, simulations, and idealizations in science, and a closer examination of actual scientific practice, has revealed that scientific understanding can be achieved without the use of traditionally-defined scientific explanations, and that the simple possession of explanatory knowledge is often not sufficient for the working scientist's understanding of a phenomenon. Scientific understanding thus seems to be a topic worth investigating in its own right.

---

[2] I will use *decision* as the general term to encompass outputs from AI models, such as predictions, categorizations, action selection, etc.

[3] Needless to say, each of these differences has been the subject of great philosophical controversy. I am simply reporting some of the reasons that have been stated in the literature to motivate the analysis of understanding as an independent concept.

There are many aspects of this literature that are germane to XAI. Here I will only focus on two main issues. The first one regards the relationship between explanation and understanding in the context of opaque machine learning models. While many authors defend the idea that there is no understanding without explanation, the impossibility of finding explanations, in the traditional sense of the term, for black box machine learning models should lead us to question the inseparability of these two concepts in the context of AI. The literature suggests alternative paths to achieve understanding, and it is worth investigating how these paths can be fruitfully adapted to understand opaque models and decisions.

The second issue regards the nature of understanding itself. Are understanding the decision of a model and understanding the model that produced that decision two states that demand different accounts or can they be reduced to the same underlying cognitive processes and abilities? I will argue that post-hoc interpretability and model transparency correspond to different levels of the same type of understanding. There is, however, a different kind of understanding that stems from the functional or instrumental analysis of machine learning models. I will argue that functional understanding falls short in many respects of the stated goals of XAI.

It should be noted that the notion of *opacity* in machine learning is itself in need of further specification. There are many types of machine learning models that are purposely designed as black boxes (e.g. deep neural networks and Support Vector Machines). Other methods, such as rule lists, linear regressions, simple naïve Bayes classifiers, and decision trees are often interpretable, but not always. "Sufficiently high dimensional models, unwieldy rule lists, and deep decision trees could all be considered less transparent than comparatively compact neural networks" (Lipton 2016, p. 5). Other relatively simple models will be opaque only to certain users who lack the required background knowledge to understand them. To simplify the object of analysis, in this paper I will only focus on the extreme case of models that are unambiguously designed as black boxes. Most of the results of this analysis can then be naturally extended to models and methods that are opaque only in certain cases or to specific stakeholders.

Finally, given the variety of purposes of black box machine learning models, and the differences in background knowledge and interests of their stakeholders, there is no reason to believe that a single interpretative strategy will be equally successful in all cases. Designing interpretative models and tools will inevitably require taking into account the psychological and pragmatic aspects of explanation. The study of the cognitive aspects of interpretative models is in its infancy. It follows from the general outlook that I present in this paper that this area of research should receive more attention in the coming years.

The essay is organized as follows. The next section examines the question of whether there are independent theoretical reasons to appeal to the notion of understanding in XAI or whether it will be sufficient, as the traditionalists claim, to develop the best possible account of explanation in AI and let understanding naturally emerge from it. I will argue that the connection between explanation and understanding in AI is not comparable to that same connection in the natural and social sciences. The disconnection arises from the impossibility, in most cases, to offer an explanation that fulfills the factivity condition. This will lead, in section 3, to a discussion about alternative paths to understanding that are not based on traditional explanations. I show how these alternative paths are exemplified in some recent attempts to find adequate methods and devices to understand opaque models and decisions. In section 4, I analyze the types of understanding that emerge from these different avenues to understanding. This will require introducing a distinction between understanding-why, which prima facie is the type of understanding involved in post-hoc interpretability, and objectual understanding, which requires grasping the inner workings of a complex system such as an AI model. This section also addresses the functional understanding of AI systems. Using evidence from psychology, it will be possible to offer a nuanced analysis of the interconnections between these three possible ways of characterizing understanding in AI.

## 2. Why not settle for AI-explanations?

A great number of philosophers of science have argued that understanding is inextricably linked to explanation. For Salmon, a defender of the ontic conception of explanation, "understanding results from our ability to fashion scientific explanations" (1984, p. 259). In more recent times, Strevens has staunchly defended the idea that "explanation is essentially involved in scientific understanding" (2013, p. 510). Perhaps the strongest claim in this direction is made by Khalifa, who defends the reductionist thesis that "any philosophically relevant ideas about scientific understanding can be captured by philosophical ideas about the epistemology of scientific explanation without loss" (2012, p. 17). In the context of XAI, this thesis[4] implies that understanding an AI model or decision is simply a question of finding an adequate explanation for it. But the implication holds only if scientific explanations and AI-explanations share a sufficient number of essential characteristics to be considered two species of the same genus. If they are, our task will be to find in AI-explanations the same features that enable scientific explanations to generate understanding. However, in this section I will argue that explanations in the present stage of AI are incommensurable with the types of explanations discussed in the philosophy of science.

My first task will be to clarify what I mean by an AI-explanation. The notion of explanation in what is often referred to as "Good Old-Fashioned AI" (GOFAI), that is, in symbolic, logic-based AI models, differs significantly from the present task of explaining opaque machine learning models. The function of an explanation in a logic-based system, either monotonic or nonmonotonic, is to support the addition of an input to a belief set or a database. For example, an update request "insert ($\varphi$)" can be achieved by finding some formula consistent with the database such that the union of the set of ground facts in the database and the formula yields $\varphi$ as a logical consequence. In previous work (Páez 2009) I argued that this abductive task is at odds with the way in which explanation has historically been understood in the philosophy of science. I refer the reader to the paper for the relevant arguments. My purpose here is to defend the same conclusion for the notion of explanation as it is being used in the field of computational intelligence in recent times.

---

[4] Here I will not evaluate the merits of this thesis in the philosophy of science. For a discussion, see the collection edited by De Regt, Leonelli and Eigner (2009).

An important difference between explanation in logic-based models and in current machine learning models is that the *explanandum* is entirely different. In the former, as just mentioned, the goal is to justify an input. In the latter, it is to explain an output, generally a decision, or to provide explanatory information about the workings of the model that generated that output. The *explanandum* of an AI-explanation as it is currently conceived is thus similar to the outcome of a scientific experiment, or to the structure of a physical or social system. A natural scientist and the stakeholder of a machine learning model would thus be searching for explanations for similar objects. But that is as far as the similarities go. In what follows I will present three fundamental reasons why it is misguided to make our understanding of machine learning models dependent on establishing an account of AI-explanations, even if we were to accept the claim that scientific understanding depends on devising bona fide scientific explanations.

The first reason has to do with truth. An essential feature of explanations in science is their factivity (Hempel 1965), i.e., both the *explanans* and the *explanandum* must be true.[5] If one denies the factivity of explanation, the claim goes, one cannot avoid the conclusion that the Ptolemaic theory, the phlogiston theory, or the caloric theory, provided bona fide scientific explanations. An explanation of *x* must reveal, depending on which theory of explanation one adopts, either the true causal structure of *x* or the natural laws that determine *x* or its relationship with factors that make *x* more or less probable.[6] All objectivist theories of explanation assume that researchers have epistemic access either to the inner workings of *x* or to the complete[7] causal or probabilistic context that determines the properties of the

---

[5] More precisely, the explanans-statement and the explanandum-statement must be true. If one holds, following Lewis (1986) and Woodward (2003), that the relata of the explanation relation are particulars, i.e., things or events, the claim amounts to saying that the things or events occurring in both the explanans and the explanandum position exist or occur.

[6] This list is not meant to be exhaustive and it excludes pragmatic theories of explanation such as the ones defended by Achinstein (1983) and van Fraassen (1980). I have argued elsewhere (Páez 2006) that these theories offer an account of explanation that lacks any sort of objectivity.

[7] Salmon's (1971) reference class rule, for example, requires the probabilistic (causal) context of a single event to be complete to avoid any epistemic relativity.

*explanandum*. Without such epistemic access it would be impossible to reach true explanatory information about *x*.

This kind of epistemic access is blocked in the case of opaque AI models. A general knowledge of the structure of a deep neural network will be insufficient to explain, in this traditional sense, either a specific decision or the actual computation that was made to generate it. Many types of black box models, like deep neural networks, are stochastic (non-deterministic). Randomness is introduced in data selection, training, and evaluation to help the learning algorithm be more robust and accurate.[8] Examining the training set and all the weights, biases and structure of the network will not allow us to understand its specific decisions, and its predictive failures and successes cannot be traced back to particular causal paths in its hidden layers. To be sure, it is possible to give a true explanation of the general design and purpose of a black box model, but such an explanation will not be sufficient to explain specific decisions or to generate trust in the model.

One of the main virtues of replacing explanation by understanding as the focus of analysis in XAI is that the factivity condition need not be satisfied. According to so-called moderate factivists (Kvanvig 2009b; Mizrahi 2012; Carter & Gordon 2016), not all the information on the basis of which something is understood must be true, only the central propositions. Other philosophers go even further and reject the factivity condition altogether.[9] Elgin's (2007) discussion of the role of models and idealizations allows that our understanding of some aspects of reality may be literally false. Far from being an unfortunate expedient, idealizations and models are an essential and ineliminable component of our scientific understanding of the world; she calls them "felicitous falsehoods" (2004, p. 116). In section 3 I will explore Elgin's idea in the context of our understanding of opaque models. I will argue that although the methods and artifacts used to understand an intelligent system and its decisions are not, and perhaps cannot be, entirely faithful to the model, this does not tell against them. On the contrary, they can afford indirect epistemic access to matters of fact that are otherwise humanly impossible to discern.

---

[8] I am grateful to an anonymous reviewer for pointing this out.
[9] The relaxation of the factivity condition is often defended in the context of objectual understanding, but it remains controversial in the case of understanding why. I return to this distinction in section 4.

A second reason to shift our focus from explanation to understanding is the importance of taking into account the specific context, background knowledge, and interests of end-users and stakeholders of opaque models.[10] In any field it is possible to establish a distinction between different levels of expertise and different levels of understanding depending on the depth of a person's knowledge of a phenomenon. In the sciences, it is expected that the novice will become an expert by acquiring the required knowledge and skills. More importantly, scientific experts will be able to master the best possible explanations of the phenomena within their field of study. This situation is not replicated in the case of machine learning. The medical doctor or the parole officer who makes use of a black box model is not supposed to acquire the level of expertise of a computer scientist, and their respective level of understanding of any explanatory model of the opaque system will remain incomparable. This seems to be an element that has not always been kept in mind in XAI. Many AI researchers build explanatory models for themselves, rather than for the intended users, a phenomenon that Miller et al. (2017) refer to as "the inmates running the asylum" (p. 36). The alternative they propose, and which I fully endorse, is to incorporate results from psychology and philosophy to XAI.[11] It is necessary to explore a naturalistic approach to the way in which context and background knowledge mold an agent's understanding of an interpretative model. Existing theories of how people formulate questions and how they select and evaluate answers should also inform the discussion (Miller 2019).

A third advantage of focusing on the pragmatic elements of interpretative models is that we can obtain a better grasp of the relationship between explanation and trust. When using machine learning in high-stakes contexts such as medical diagnosis or parole decisions it is necessary to trust the individual decisions generated by the model. Several authors have argued that post-hoc interpretability, i.e., an explanation of the decision, is a necessary condition for trust (Kim 2015, Ribeiro et al. 2016). Additionally, of course, the system must have a very high score on an evaluation metric based on decisions and ground truths. Suppose that an opaque model has consistently shown a high degree of predictive accuracy and a user has been given a clear post-hoc explanation of its behavior. The user has the best possible understanding of the system, taking into account, of course, the epistemic limitations mentioned above. But predictive reliability and a post-hoc explanation are not sufficient to generate trust. Trust does not depend exclusively on epistemic factors; it also depends on the interests, goals, resources, and degree of risk aversion of the stakeholders. Trust involves a *decision* to accept an output and *act* upon it. Different agents bound by different contextual factors will make different decisions on the basis of the same information. I will leave open the question of whether classical decision theory can provide an adequate analysis of trust in AI systems.[12] But the important lesson to draw from the multidimensional character of trust is that there is no simple correlation between explanation and trust, and that an adequate analysis of trust requires taking into account contextual factors that can foster or hinder it.

---

[10] De Graaf and Malle (2017) have also emphasized the importance of these pragmatic factors: "The success of an explanation therefore depends on several critical audience factors—assumptions, knowledge, and interests that an audience has when decoding the explanation" (p. 19).
[11] See also De Graaf & Malle (2017), Miller (2019), and Mittelstadt et al. (2019).

The reasons I have presented in this section recommend abandoning the traditional "explanationist" path according to which understanding can only be obtained via an explanation in any of the guises it has adopted in the philosophy of science. The next section will offer alternative ways to achieve understanding.

## 3. Alternative Paths to Understanding

Abandoning the necessary connection between explanation and understanding opens up several avenues of research that can lead to understanding the workings and decisions of opaque models. Implicit causal knowledge, analogical reasoning, and exemplars are obvious alternative paths to understanding. But so are models, idealizations, simulations, and thought experiments, which play important roles in scientific understanding despite being literally false representations of their objects. In a similar vein, the methods and devices used to make black box models understandable need not be propositionally-stated explanations and they need not be truthful representations of the models. I will begin by presenting a few examples of how understanding can be achieved in the natural sciences without the use of explanations before moving to a discussion of how similar devices can be used, and have been used, in understanding AI models.

---

[12] See Falcone & Castelfranchi (2001) for a critique of the use of decision theory to understand trust in virtual environments.

Many philosophers, beginning with Aristotle and continuing with the defenders of causal explanations, have argued that understanding-why is simply knowledge of causes (Salmon 1984; Lewis 1986; Greco 2010; Grimm 2006, 2014).[13] Naturally, causal explanations are the prime providers of knowledge of causes. But causal knowledge does not come exclusively from explanation. As Lipton points out, "much of empirical inquiry consists in activities–physical and intellectual–that generate causal information, activities such as observation, experimentation, manipulation, and inference. And these activities are distinct from the activity of giving and receiving explanations" (2009, p. 44). To be sure, the causal information generated by these activities can be given a propositional representation and can thus be transformed into explicit causal explanations. But Lipton argues that in many cases such activities generate causal information that remains as tacit knowledge, allowing us to perform epistemic and practical tasks. Such tacit causal knowledge comes primarily from images and physical models. An orrery or a video, for example, can provide better understanding of retrograde planetary motion than an explanation stated in propositional form. A subject might even be able to understand retrograde motion without being able to articulate such an explanation.

Direct manipulation or tinkering of a causal system is an even more obvious source of implicit causal knowledge. Adjusting a lever, a button or an input variable and observing its effects on other parts of a system is a way of beginning to understand how the system works. Manipulation also provides modal information about the possible states of a system. In fact, the ability to manipulate a system into new desired states should be seen as a sign of understanding. In other words, understanding requires the ability to think counterfactually (de Regt & Dieks 2005; Wilkenfeld 2013).

---

[13] Philosophers of science are much more inclined to accept this view than epistemologists, who have fiercely resisted it. See, for example, Zagzebski (2001), Kvanvig (2003), Elgin (2004), and Pritchard (2014). I do not have space to discuss the issue here, but from the text it should be clear that I side with the epistemologists.

Causal information, implicit or explicit, is not the only source of understanding. Consider analogical reasoning. Darwin (1860/1903) used an analogy between the domestic selection of animals and natural selection to argue for the latter. Although it is incomparable in many respects, artificial selection illuminates how the mechanism would work in a larger class (Lipton 2009, p. 51). Exemplification is another important avenue towards understanding. The examples in a logic textbook can show a student how the rules of natural deduction work. Her initial understanding of the rules will be tied to the examples, but it will gradually drift away as her ability to use the rules in new situations improves. When an item serves as an example, "it functions as a symbol that makes reference to some of the properties, patterns, or relations it instantiates" (Elgin 2017, p. 184). It can only display some of these features, downplaying or ignoring others. As the complexity of the item increases, the decision to emphasize or underscore some salient features over others will be determined by pragmatic reasons, such as the intended audience and use of the example.

The use of non-propositional representations such as diagrams, graphs, and maps present another clear case of understanding without explanation. A subway map is never a faithful representation of the real train network. It alters the distance between stations and the exact location of the tunnels in order to make the network easy to understand, but it must include the correct number of lines, stations and intersections to be useful at all. It must be sufficiently accurate without being too accurate.

Finally, models and idealizations play a similar role in science (Potochnik 2017). They simplify complex phenomena and sometimes the same phenomenon is represented by multiple, seemingly incongruous models. They afford epistemic access to features of the object that are otherwise difficult or impossible to discern. Models are not supposed to accurately represent the facts, but they must be objective. Models have to *denote* in some sense the facts they model. They "are representations *of* the things that they denote" (Elgin 2008, p. 77). The general relation between scientific models and their objects is a thorny issue that deserves a more detailed discussion than the one I can provide here, but one important aspect that must be noted is that the adequate level of "fit" between a model and its object is a pragmatic question. Many models are, in Elgin's apt phrase, "true enough" of the phenomenon they denote:

This may be because the models are approximately true, or because they diverge from truth in irrelevant respects, or because the range of cases for which they are not true is a range of cases we do not care about, as for example when the model is inaccurate at the limit. Where a model is true enough, we do not go wrong if we think of the phenomena as displaying the features exemplified in the model. Obviously, whether such a representation is true enough is a contextual question. A representation that is true enough for some purposes, or in some respects is not true enough for or in others (2008, p. 85).

Applications of all of the approaches mentioned above can be found in the XAI literature. It is important to bear in mind that many authors in the field refer to these alternative paths to understanding as "explanations," a usage that threatens to trivialize the term. If whatever makes an opaque model or its decisions better understood is called an explanation, the term ceases to have any definitive meaning. My argument throughout the paper has only focused on the notion of explanation as it has been traditionally understood in the philosophy of science and epistemology (e.g., causal models, covering-law models, probabilistic approaches, etc.). It is in this sense that there are alternative sources of understanding.

It is customary to distinguish between two different goals in XAI: understanding a decision, often called post-hoc interpretability, and understanding how the model functions, i.e., making the model transparent (Lipton 2016; Lepri et al. 2017; Mittelstadt et al. 2019). Exemplifications, analogies, and causal manipulation are often used in the former, while the use of models is more common in the latter. I will present some examples of the use of these techniques, and in the next section I will examine the kind of understanding they provide. The ultimate question I will try to answer is whether transparency and post-hoc interpretability aim at different types of understanding.

The attempts to make a model transparent can focus on the model as a whole (simulatability), on its parameters (decomposability), or on its algorithms (algorithmic transparency) (Lipton 2016). A complete understanding of the model would thus allow a user to repeat (simulate) the computation process with a full understanding of the algorithm and an intuitive grasp of every part of the model. Each of these aspects presents its own challenges, but my interest here is in the use of interpretative devices to provide an overall understanding of opaque models, i.e., models that are not designed to be fully understood. The most common way to make a black-box model as understandable as possible is through the use of proxy or interpretative models (Guidotti et al. 2018). Many of these models provide coarse approximations of how the system behaves over a restricted domain. The two most widely used classes of models are linear or gradient-based approximations, and decision trees (Mittelstadt et al. 2019). For the interpretative model to be useful, a user must know "over which domain a model is reliable and accurate, where it breaks down, and where its behavior is uncertain. If the recipient of a local approximation does not understand its limitations, at best it is not comprehensible, and at worst misleading" (Mittelstadt et al. 2019, p. 281). Oversimplified or misleading models also incur the risk of being perceived as deceitful, thereby undermining the user's trust in the original model. Thus, the first desideratum of interpretative models is that they must be as faithful to the original model as possible and absolutely transparent about their limitations.

Mittelstadt et al. (2019) argue that XAI should not focus on developing interpretative models because they are akin to scientific models, and therefore very different from "the types of scientific and 'everyday' explanations considered in philosophy, cognitive science, and psychology" (p. 279). My view is exactly the opposite. Since the notion of explanation discussed in the philosophy of science is inapplicable in the context of opaque machine learning models, and since I do not want to settle for a purely subjective sense of explanation, XAI should adopt any other methods and devices that provide objective understanding. Scientific models, suitably adapted to the intended users, offer an indirect[14] path towards an objective understanding of a phenomenon. We should therefore see the parallel between scientific models and interpretative models in a positive light.

The fidelity desideratum for interpretative models has to be balanced against the desideratum of comprehensibility. There are very few empirical studies about which kinds of interpretative models are easier to understand. Huysmans et al. (2011), for example, present evidence that single-hit decision tables perform better than binary decision trees, propositional rules, and oblique rules in terms of accuracy, response time, and answer confidence for a set of problem-solving tasks involving credit scoring. This study is of limited use because it was done with extremely simple representation formats and the only participants were 51 graduate business students.[15] It is necessary to undertake similar studies that also include linear regressions, simple naïve Bayes classifiers, and random forests.[16] These interpretative models also have to be tested on a more diverse population with different levels of expertise (Doshi-Velez & Kim 2017). These types of empirical studies are essential for the purposes of XAI, and they have to be complemented with psychological studies of the formal and contextual factors that enhance understanding. As noted by Pazzani (2000), there is little understanding of the factors that foster or hinder interpretability in these cases, and of whether users prefer, for example, visualizations over textual representations.

---

[14] A *direct* understanding of a phenomenon would be factive, based on a literal description of the explanatory elements involved. It is in this sense that models offer an *indirect* path towards objective understanding.

The appropriateness of an interpretative model thus depends on three factors: obtaining the right fit between the interpretative model and the black box model in terms of accuracy and reliability, providing sufficient information about its limitations, and achieving an acceptable degree of comprehensibility for the intended user. While there may be some identifiable, permanent features of interpretative models that facilitate understanding, the choice of the best proxy method or artifact will also depend on who the intended users of the original system are. Their background knowledge, their levels of expertise, and the time available to them to understand the proxy model can vary widely. This last aspect has been entirely neglected in the literature; not a single method reviewed by Guidotti et al. (2018) presents real experiments about the time required to understand an interpretative model.

---

[15] In Allahyari and Lavesson (2011), 100 non-expert users were asked to compare the understandability of decision trees and rule lists. The former method was deemed more understandable. Freitas (2014) examines the pros and cons of decision trees, classification rules, decision tables, nearest neighbors, and Bayesian network classifiers with respect to their interpretability, and discusses how to improve the comprehensibility of classification models in general. More recently, Fürnkranz et al. (2018) performed an experiment with 390 participants to question the idea that the likeliness that a user will accept a logical model such as rule sets as an explanation for a decision is determined by the simplicity of the model. Lage et al. (2019) also explore the complexities of rule sets to find features that make them more interpretable, while Piltaver et al. (2016) undertake a similar analysis in the case of classification trees. Another important aspect of this empirical line of research is the study of cognitive biases in the understanding of interpretable models. Kliegr et al. (2018) study the possible effects of biases on symbolic machine learning models.

[16] As noted in the Introduction, none of these methods is *intrinsically* interpretable.

Turning very briefly to post-hoc interpretability, we find in the literature several interpretative devices to understand a decision. In many cases, a sensitivity analysis provides a local, feature-specific, linear approximation of the model's response. The result of the analysis consists of a list, a table, or a graphical representation of the main features that influenced a decision and their relative importance. Often, such devices allow a certain degree of causal manipulation that brings out feature interactions. This is the basis of the LIME model proposed by Ribeiro et al. (2016), a technique to offer functional explanations of the decisions of any machine learning classifier. To understand the behavior of the underlying model, the input is perturbed to see how the decisions change without worrying about the actual computation that produced it. The user can ask counterfactual questions about local changes and see the results in an intuitive way. Saliency maps offer a similar functional understanding of the model. A network is repeatedly tested with portions of the input occluded to create a map showing which parts of the data actually have influence on the network output (Zeiler & Fergus 2014; Lapuschkin et al. 2019).

Caruana et al. (1999) argue that analogies and exemplars (prototypes) are a useful heuristic device. A model can report, for every new decision, other examples in the training set that the model considers to be most similar. The authors seek to use this method in clinical contexts, where doctors often refer to case-studies to justify a course of action. The basic assumption made by case-based methods, such as *k*-nearest neighbor, is that similar inputs correlate with similar outputs. The methods look for the case in the training set, the prototype, that is most similar in terms of input features to the case under consideration.

Another commonly used method, especially in interactions with autonomous agents, is to provide natural language explanations of a decision (McAuley & Leskovec 2013; Krening et al. 2016). These explanations state information about the most important features in a decision and come closer than any other method to the causal explanations used in science and everyday life. The difference, once again, is that these "explanations" are not factive, regardless of how plausible they appear. Eshan et al. (2018) even suggest that textual explanations can be rationalizations: "AI rationalization is based on the observation that there are times when humans may not have full conscious access to reasons for their behavior and consequently may not give explanations that literally reveal how a decision was made. In these situations, it is more likely that humans create plausible explanations on the spot when pressed. However, we accept human-generated rationalizations as providing some lay insight into the mind of the other" (p. 81).

A common feature of many post-hoc interpretations is that they are model-agnostic. They do not even attempt to open the black box and they offer only a functional approach to the problem of explaining a decision. The cognitive achievement reached by the use of these devices seems to differ in great measure from the understanding provided by an interpretative model. In the last section of the paper I will tackle the question of whether it is possible to characterize different types of understanding in AI.

## 4. Types of Understanding in AI

On the basis of the methods described in the previous section, it is tempting to divide the understanding they provide into two different types. The first one would be associated with post-hoc interpretability. This type is often called *understanding-why*, and in this case its object will be a specific decision of a model. In contrast, transparency seems to generate an *objectual understanding* of a model. The distinction between these two types of understanding has been widely discussed in epistemology. The question I will examine in the beginning of this section is whether this epistemological distinction can be defended in the present context.

Epistemologists establish a distinction between understanding why something is the case, and understanding an object, a system or a body of knowledge (Kvanvig, 2003). It seems straightforward to say that the goal of transparency in machine learning can be understood in terms of objectual understanding. Consider the various ways in which this type of understanding has been described: According to Zagzebski, understanding "involves grasping relations of parts to other parts and perhaps the relation of parts to a whole" (2009, p. 144). For Grimm, the target of objectual understanding is a "system or structure … that has parts or elements that depend upon one another in various ways" (2011, p. 86). And Greco characterizes it as "knowledge of a system of dependence relations" (2012, p. 123). The interpretative models that we considered in the previous section all provide the kind of understanding described by these authors.

Understanding why *p*, on the other hand, is not equivalent to simply *knowing* why *p*. Suppose the only thing a person knows about global warming is that it is caused, to a large extent, by an increase in the concentration of greenhouse gases. This is a claim the person has heard repeatedly in serious media outlets and scientific TV shows, but he has never stopped to think about the causal mechanisms involved. The person *knows* why the earth is warming, but this information is insufficient to *understand* why it is warming. The person lacks, for example, the ability to answer a wide range of questions of the type what-if-things-had-been-different (Woodward 2003, p. 221). What would happen to global temperatures if all human activity were to cease? What would be the effect on global warming of a massive volcanic eruption similar in scale to the eruption of Krakatoa in 1883? These are the kind of counterfactual scenarios commonly studied in climate research and modelling, which the common person is unable to understand. A complete understanding of global warming also involves the ability to make probability estimates of future scenarios based on current data.

Notice that the ability to answer counterfactual questions and to make predictions depends to a large extent on an objectual understanding of the larger body of knowledge to which the specific object of understanding belongs. Without a basic understanding of the structure, chemistry, and behavior of the earth's atmosphere, for example, a person will not be able to answer counterfactual questions or deliver probability estimates about global warming. It follows, as Grimm (2011) convincingly argues, that understanding-why is a variety of objectual understanding, but at a local level, and that there is no genuine distinction between the two types of understanding. The implication for machine learning is that understanding a decision requires some degree of objectual understanding of the model. Mittelstadt et al. (2019) seem to reach a similar conclusion:

> [A]t the moment, XAI generally avoids the challenges of testing and validating approximation models, or fully characterizing their domain. If these elements are well understood by the individual, models can offer more information than an explanation of a single decision or event. Over the domain for which the model accurately maps onto the phenomena we are interested in, it can be used to answer 'what if' questions, for example "What would the outcome be if the data looked like this instead?" and to search for contrastive explanations, for example "How could I alter the data to get outcome X?" (p. 282).[17]

It is true that some of the post-hoc interpretability devices described in the previous section allow stakeholders to manipulate the parameters and observe the different decisions generated thereby. But this is not genuine counterfactual reasoning. By tinkering with the parameters, the stakeholders can only form functional generalizations with a very weak inductive base. True counterfactual reasoning is purely theoretical, based on knowledge about how the model works. Thus, if we take the ability to think counterfactually about a phenomenon as a sign that the agent understands it, as suggested by de Regt and Dieks (2005), understanding the decisions of a model requires some degree of objectual understanding.

There is, nonetheless, an important difference between the two types of understanding under consideration. Virtually all epistemologists regard understanding-why as factive, while allowing that objectual understanding might not be entirely so. Pritchard, for example, gives the following example to show that understanding-why is factive: "Suppose that I believe that my house has burned down because of an act of vandalism, when it was in fact caused by faulty wiring. Do I understand why my house burned down? Clearly not" (2008, p. 8). In other words, according to Pritchard, without a true causal explanation there can be no understanding-why. But changing the example can debilitate the intuitions that support this conclusion. Suppose an engineer is investigating the collapse of a bridge and uses Newtonian physics as the basis for his analysis. Strictly speaking, the explanation is based on a false theory, but it can hardly be argued that the engineer is a priori barred from understanding why the bridge collapsed. Or suppose an economist successfully explains a sudden rise in inflation using a macroeconomic model that, again, cannot be literally true (Reiss 2012). It thus seems that the factivity of understanding-why can only be defended in simple scenarios where a complete analysis of the relevant causal variables can be provided, but as soon as the context requires the use of theoretical tools such as idealizations and models, it becomes highly doubtful.

---

[17] A terminological clarification is in order. Mittelstadt et al. (2019) and other researchers in XAI use the phrase "contrastive explanations" to refer to counterfactuals. But these are two very different things. In philosophy, an explanation is contrastive if it answers the question "Why $p$ rather than $q$?" instead of just "Why $p$?" In either case the explanation provided must be factual. To turn it into a counterfactual situation, the question must be changed to: "What changes in the world would have brought about $q$ instead of $p$?" And the answer will be a hypothetical or counterfactual statement, not an explanation.

Machine learning is precisely this kind of context. The use of arbitrary black-box functions to make decisions in machine learning makes it impossible to reach the causal knowledge necessary to provide a true causal explanation. The functions may be extremely complex and have an internal state composed of millions of interdependent values. Machine learning is the kind of context in which one can say that, in principle, it is impossible to satisfy the factivity condition for understanding-why.

We thus have an argument to the effect that understanding-why and objectual understanding in machine learning cannot be entirely independent of each other, but rather, that the former is a localized variety of the latter. And we have an argument against the claim that understanding-why is always factive, which was supposed to be the most important property that distinguished both types of understanding. So even if prima facie the devices and methods used to provide transparency and post-hoc interpretability are different, it is safe to say, on the one hand, that understanding-why and objectual understanding are two different species of the same genus, and on the other, that there is no essential difference between them in terms of truth.

There is, however, a third way of characterizing understanding in AI. Psychologists distinguish between the functional and the mechanistic understanding of an event. The former "relies on an appreciation for functions, goals, and purpose" while the latter "relies on an appreciation of parts, processes, and proximate causal mechanisms" (Lombrozo & Wilkenfeld forthcoming, p. 1). For example, an alarm clock beeps because the circuit connecting the buzzer to a power source has been completed (mechanical understanding) and because its owner has set it to wake her up at a specific time (functional understanding).

Lombrozo and Wilkenfeld argue that a subject can have a functional understanding of an event while being insensitive to mechanistic information. Lombrozo and Gwynne (2014) have shown that properties that are understood functionally, as opposed to mechanistically, are more likely to be generalized on the basis of shared functions. This means that a functional, as opposed to a mechanistic understanding of the relation between an input and an output will make it easier for a user to inductively conclude that similar inputs produce similar decisions. There is also evidence that functional reasoning may be psychologically privileged in the sense that it is often favored and seems to be less cognitively demanding than mechanistic reasoning. Humans are "promiscuously teleological," to use Kelemen's (1999) apt description. Finally, Lombrozo and Wilkenfeld also argue that functional and mechanistic understanding differ with regard to normative considerations. A functional understanding of a property of an object tells us what it is supposed to do, while understanding the mechanism that causes that property lacks this normative element. Functional understanding thus seems to be a different kind of understanding altogether, compared to objectual understanding and understanding-why. It involves different content, it supports different functions, and it has a distinctive phenomenology.

If we take the decision of an opaque model as our object of understanding, a mechanistic understanding of it is equivalent to the local objectual understanding of the model, as I have argued above. Its functional understanding, on the other hand, would focus on the purpose of the model and the relation between its features and decisions. Functional reasoning about black box models allows for a more mechanism-independent form of reasoning. Aiming at this type of understanding will be appealing to those who want to offer model-neutral interpretability devices and focus only on covariations between inputs and decisions.

However, it seems to me that aiming for functional understanding in XAI is, to a certain extent, to give up on the project of explaining why an AI model does what it does. It is to embrace the black box and trust it as one trusts a reliable oracle without understanding its mysterious ways. Less metaphorically, reliability by itself cannot usher trust because of the dataset shift problem (Quinonero-Candela et al. 2009). To have confidence that the model is really capturing the correct patterns in the target domain, and not just patterns valid in past data that will not be valid in future data, it is necessary to have a global or at least a local objectual understanding of the model.[18] Unfortunately, most solutions to the dataset shift problem focus only on accuracy ignoring model comprehensibility issues (Freitas 2014). Furthermore, methods designed to enhance the functional understanding of a model are also more likely to be tailored to user preferences and expectations, and thus prone to oversimplification and bias. Although the understanding and trust sought by XAI should always take into account a model's stakeholders, it should not pursue these goals by offering misleadingly simple functional explanations that can derive in unjustified or dangerous actions (Gilpin et al. 2019). Finally, a purely functional understanding of a model would also impede legal accountability and public responsibility for the decisions of the model. Guilt for an unexpected decision with harmful or detrimental consequences to the user cannot be decided if the only information available is the previous predictive accuracy of the model. It is necessary to understand why the model produced the unexpected result, that is, to have a local objectual understanding of it.

In sum, in this section I have argued that both transparency and post-hoc interpretability should be seen as more or less encompassing varieties of objectual understanding, and that the kind of understanding provided by the functional approach to a model offers an understanding of a different and more limited kind. In my view, it is the former kind that should interest researchers in XAI.

## 5. Conclusion

In this paper I have argued that the term 'explanation', as it is currently used in XAI, has no definitive meaning and shares none of the properties that have been traditionally attributed to explanations in epistemology and the philosophy of science. My suggestion has been to shift our focus from a blind search for explanatory devices and methods whose success is uncertain, to the study of the mental state that XAI researchers are aiming at, namely, an objective understanding of opaque machine learning models and their decisions. I have argued that the use of interpretative models is the best avenue available to obtain understanding, both in terms of transparency (understanding how the model works) and post-hoc interpretability (understanding a decision of the model). The current approaches to the latter rely on a purely functional understanding of models; however, leaving the black box entirely untouched seems to belie the purpose of XAI. It must be admitted that interpretative models can provide false assurances of comprehensibility. The task ahead for XAI is thus to fulfill the double desiderata of finding the right fit between the interpretative and the black box model, and to design interpretative models and devices that are easily understood by the intended users. This latter task must be guided by an empirical investigation of the features of interpretative models that make them easier to understand to users with different backgrounds and levels of expertise. One of the possible areas of research is the comparative study of the complexity of rule sets, decision tables and trees, nearest neighbors, and Bayesian network classifiers with respect to their interpretability. XAI can also benefit from interdisciplinary work with designers to create user-friendly, accessible, and engaging interpretative tools and interfaces, in the same spirit as the legal design movement. Finally, an important aspect of this empirical line of research is the study of cognitive biases in the interpretation of models, especially in the context of autonomous systems with human-like interfaces.

---

[18] To be sure, there are many scenarios where both the owner and the user (but not the developer) of the model will be satisfied with its accurate decisions without feeling the need to have an objectual understand of it. Think of the books recommended by Amazon or the movies suggested by Netflix using the simple rule: "If you liked *x*, you might like *y*." As I argued in section 2, the relation between understanding and trust is always mediated by the interests, goals, resources, and degree of risk aversion of stakeholders. In these cases, the cost-benefit relation makes it unnecessary to make the additional effort of looking for mechanisms.

## References

Achinstein, P. (1983). *The nature of explanation*. New York: Oxford University Press.

Allahyari, H., & Lavesson, N. (2011). User-oriented assessment of classification model understandability. *Proceedings of the 11th Scandinavian Conference on Artificial Intelligence*. Amsterdam: IOS Press.

Carter, J. A., & Gordon, E. C. (2016). Objectual understanding, factivity and belief. In: M. Grajner & P. Schmechtig (Eds.), *Epistemic reasons, norms and goals* (pp. 423-442). Berlin: De Gruyter.

Caruana, R., Kangarloo, H., Dionisio, J. D. N., Sinha, U., & Johnson, D. (1999). Case-based explanations of non-case-based learning methods. In *Proceedings of the AMIA Symposium* (p. 212). American Medical Informatics Association.

Castelfranchi, C., & Tan, Y-H. (Eds.). (2001). *Trust and deception in virtual societies* (pp. 157-168). Dordrecht: Kluwer Academic Publisher.

Darwin, C. (1860/1903). Letter to Henslow, May 1860. In F. Darwin (Ed.), *More letters of Charles Darwin, vol. I*. New York: D. Appleton.

De Graaf, M. M., & Malle, B. F. (2017). How people explain action (and Autonomous Intelligent Systems should too). In *AAAI Fall Symposium on Artificial Intelligence for Human-Robot Interaction* (pp. 19-26). Palo Alto: The AAAI Press.

de Regt, H. W., & Dieks, D. (2005). A contextual approach to scientific understanding. *Synthese*, *144*, 137–170.

de Regt, H. W., Leonelli, S., & Eigner, K. (Eds.). (2009). *Scientific understanding: Philosophical perspectives*. Pittsburgh: University of Pittsburgh Press.

Doshi-Velez, F., & Kim, B. (2017). Towards a rigorous science of interpretable machine learning. *arXiv preprint arXiv:1702.08608*.

Ehsan, U., Harrison, B., Chan, L., & Riedl, M. O. (2018). Rationalization: A neural machine translation approach to generating natural language explanations. In *Proceedings of the 2018 AAAI/ACM Conference on AI, Ethics, and Society* (pp. 81-87). New York: ACM.

Elgin, C. Z. (2004). True enough. *Philosophical Issues*, *14*, 113-131

Elgin, C. Z. (2007). Understanding and the facts. *Philosophical Studies*,*132*, 33–42.

Elgin, C. Z. (2008). Exemplification, idealization, and scientific understanding. In M. Suárez (Ed.), *Fictions in science: Philosophical essays on modelling and idealization* (pp. 77-90). London: Routledge.

Elgin, C. Z. (2017). *True enough*. Cambridge: MIT Press.

Falcone R., & Castelfranchi, C. (2001). Social trust: A cognitive approach. In C. Castelfranchi, & Tan, Y.-H. (Eds), *Trust and deception in virtual societies* (pp. 55-90). Springer: Dordrecht.

Freitas, A. A. (2014). Comprehensible classification models: a position paper. *ACM SIGKDD explorations newsletter*, *15(1)*, 1-10.

Fürnkranz, J., Kliegr, T., & Paulheim, H. (2018). On cognitive preferences and the plausibility of rule-based models. *arXiv preprint arXiv:1803.01316*.

Gilpin, L. H., Bau, D., Yuan, B. Z., Bajwa, A., Specter, M., & Kagal, L. (2019). Explaining explanation. An overview of interpretability of machine learning. *arXiv preprint arXiv:1806.00069v3*.

Greco, J. (2010). *Achieving knowledge*. Cambridge: Cambridge University Press.

Greco, J. (2012). Intellectual virtues and their place in philosophy. In C. Jäger & W. Löffler (Eds.), *Epistemology: Contexts, values, disagreement: Proceedings of the 34th International Wittgenstein Symposium* (pp. 117-130). Heusenstamm: Ontos.

Grimm, S. R. (2006). Is understanding a species of knowledge? *British Journal for the Philosophy of Science*, *57*, 515-535.

Grimm, S. R. (2011). Understanding. In S. Bernecker & D. Pritchard (Eds.), *The Routledge companion to epistemology* (pp. 84-94). New York: Routledge.

Grimm, S. R. (2014). Understanding as knowledge of causes. In A. Fairweather (Ed.), *Virtue epistemology naturalized: Bridges between virtue epistemology and philosophy of science*. Dordrecht: Springer.

Grimm, S. R. (Ed.). (2018). *Making sense of the world: New essays on the philosophy of understanding*. New York: Oxford University Press.

Guidotti, R., Monreale, A., Ruggieri, S., Turini, F., Giannotti, F., & Pedreschi, D. (2018). A survey of methods for explaining black box models. *ACM Computing Surveys (CSUR)*, *51(5)*, Article 93.

Hempel, C. G. (1965). *Aspects of scientific explanation*. New York: The Free Press.

Huysmans, J., Dejaeger, K., Mues, C., Vanthienen, J., & B. Baesens (2011). An empirical evaluation of the comprehensibility of decision table, tree and rule based predictive models. *Decision Support Systems*, *51(1)*, 141-154.

Kelemen, D. (1999). Functions, goals, and intentions: Children's teleological reasoning about objects. *Trends in Cognitive Science*, *12*, 461-468.

Khalifa, K. (2012). Inaugurating understanding or repackaging explanation. *Philosophy of Science*, *79*, 15-37.

Kim, B. (2015). *Interactive and interpretable machine learning models for human machine collaboration*. PhD thesis, Massachusetts Institute of Technology.

Kliegr, T., Bahník, Š., & Fürnkranz, J. (2018). A review of possible effects of cognitive biases on interpretation of rule-based machine learning models. *arXiv preprint arXiv:1804.02969*.

Krening, S., Harrison, B., Feigh, K., Isbell, C., Riedl, M., & Thomaz, A. (2016). Learning from explanations using sentiment and advice in RL. *IEEE Transactions on Cognitive and Developmental Systems*, *9(1)*, 44-55.

Kvanvig, J. (2003). *The value of knowledge and the pursuit of understanding*. New York: Cambridge University Press.

Kvanvig, J. (2009). Response to critics. In A. Haddock, A. Millar, & D. Pritchard (Eds.), *Epistemic value* (pp. 339–351). New York: Oxford University Press.

Lage, I., Chen, E., He, J., Narayanan, M., Kim, B., Gershman, S., & Doshi-Velez, F. (2019). An Evaluation of the Human-Interpretability of Explanation. *arXiv preprint arXiv:1902.00006*.

Lapuschkin, S., Wäldchen, S., Binder, A., Montavon, G., Samek, W., & Müller, K. R. (2019). Unmasking Clever Hans predictors and assessing what machines really learn. *Nature communications*, *10*(1), 1096.

Lepri, B., Oliver, N., Letouzé, E., Pentland, A., & Vinck, P. (2017). Fair, transparent, and accountable algorithmic decision-making processes: The premise, the proposed solutions, and the open challenges. *Philosophy & Technology*, *31*, 611-627.

Lewis, D. K. (1986). Causal explanation. In *Philosophical papers, vol. II* (pp. 214-240). New York: Oxford University Press.

Lipton, P. (2009). Understanding without explanation. In H. W. de Regt, S. Leonelli, & K. Eigner (Eds.), *Scientific understanding: Philosophical perspectives* (pp. 43-63). Pittsburgh: University of Pittsburgh Press.

Lipton, Z. C. (2016). The mythos of model interpretability. *arXiv preprint arXiv:1606.03490*.

Lombrozo, T. & Gwynne, N. Z. (2014). Explanation and inference: Mechanistic and functional explanations guide property generalization. *Frontiers in Human Neuroscience*, *8*, 700.

Lombrozo, T., & Wilkenfeld, D. A. (forthcoming). Mechanistic vs. functional understanding. In S. R. Grimm (Ed.), *Varieties of understanding: New perspectives from philosophy, psychology, and theology*. New York: Oxford University Press.

McAuley, J., & Leskovec, J. (2013). Hidden factors and hidden topics: understanding rating dimensions with review text. In *Proceedings of the 7th ACM conference on recommender systems* (pp. 165-172). New York: ACM.

Miller, T. (2019). Explanation in artificial intelligence: Insights from the social sciences. *Artificial Intelligence*, *267*, 1-18.

Miller, T., Howe, P., & Sonenberg, L. (2017). Explainable AI: Beware of inmates running the asylum. In *Proceedings of the IJCAI-17 Workshop on Explainable AI (XAI)* (pp. 36-42). Accessed March 10, 2019 http://www.intelligentrobots.org/files/IJCAI2017/IJCAI-17_XAI_WS_Proceedings.pdf

Mittelstadt, B., Russell, C., & Wachter, S. (2019). Explaining explanations in AI. In *Proceedings of the Conference on Fairness, Accountability, and Transparency* (pp. 279-288). New York: ACM.

Mizrahi, M. (2012). Idealizations and scientific understanding. *Philosophical Studies*, *160*, 237-252.

Páez, A. (2006). *Explanations in K. An analysis of explanation as a belief revision operation*. Oberhausen: Athena Verlag.

Páez, A. (2009). Artificial explanations: The epistemological interpretation of explanation in AI. *Synthese*, *170*, 131-146.

Pazzani, M. (2000). Knowledge discovery from data? *IEEE Intelligent System*s, *15(2)*, 10-13.

Piltaver, R., Luštrek, M., Gams, M., & Martinčić-Ipšić, S. (2016). What makes classification trees comprehensible? *Expert Systems with Applications: An International Journal*, *62*(C), 333-346.

Potochnik, A. (2017). *Idealization and the aims of science*. Chicago: University of Chicago Press.

Pritchard, D. (2008). Knowing the answer, Understanding and epistemic value. *Grazer Philosophische Studien*, *77*, 325-339.

Pritchard, D. (2014). Knowledge and understanding. In A. Fairweather (Ed.), *Virtue scientia: Bridges between virtue epistemology and philosophy of science* (pp. 315-328). Dordrecht: Springer.

Quinonero-Candela, J., Sugiyama, M., Schwaighofer, A., & Lawrence, N. D. (Eds.). (2009). *Dataset shift in machine learning*. Cambridge: MIT Press.

Reiss, J. (2012). The explanation paradox. *Journal of Economic Methodology*, *19*, 43-62.

Ribeiro, M. T., Singh, S., & Guestrin, C. (2016). "Why should I trust you?": Explaining the predictions of any classifier. In *Proceedings of the 22nd ACM SIGKDD International Conference on Knowledge Discovery and Data Mining* (pp. 1135-1144). New York: ACM.

Salmon, W. C. (1971). Statistical explanation. In W. C. Salmon (Ed.), *Statistical explanation and statistical relevance*. Pittsburgh: Pittsburgh University Press.

Salmon, W. C. (1984). *Scientific explanation and the causal structure of the world*. Princeton: Princeton University Press.

Samek, W., Wiegand, T., & Müller, K. R. (2017). Explainable artificial intelligence: Understanding, visualizing and interpreting deep learning models. *arXiv preprint arXiv:1708.08296*.

Strevens, M. (2013). No understanding without explanation. *Studies in the History and Philosophy of Science*, *44*, 510-515.

van Fraassen, B. (1980). *The scientific image*. Oxford: Clarendon Press.

Wilkenfeld, D. (2013). Understanding as representation manipulability. *Synthese*, *190*, 997–1016.

Woodward, J. (2003). *Making things happen. A theory of causal explanation*. New York: Oxford University Press.

Zagzebski, L. (2001). Recovering understanding. In M. Steup (Ed.), *Knowledge, truth, and duty: Essays on epistemic justification, responsibility, and virtue*. New York: Oxford University Press.

Zagzebski, L. (2009). *On epistemology*. Belmont: Wadsworth.

Zeiler, M. D., & Fergus, R. (2014). Visualizing and understanding convolutional networks. In *13th European Conference on Computer Vision ECCV 2014* (pp. 818-833). Cham: Springer.